\documentclass[runningheads]{llncs}

\usepackage[T1]{fontenc}
\usepackage{graphicx}
\usepackage{hyperref}
\usepackage{color}

\usepackage{amsmath}
\usepackage{booktabs}
\usepackage{multirow}
\usepackage{todonotes}
\usepackage{cleveref}
\usepackage{amssymb}
\usepackage{subcaption}

\makeatletter
\newcommand{\printfnsymbol}[1]{%
  \textsuperscript{\@fnsymbol{#1}}%
}

\begin{document}

\title{Estimating Earthquake Magnitude in \\Sentinel-1 Imagery via Ranking}

\author{
Daniele Rege Cambrin\inst{1}\thanks{equal contribution}
\orcidID{0000-0002-5067-2118} \and \\
Isaac Corley\inst{2}\printfnsymbol{1}\orcidID{0000-0002-9273-7303} \and \\
Paolo Garza\inst{1}\orcidID{0000-0002-1263-7522} \and \\
Peyman Najafirad\inst{2}\orcidID{0000-0001-9671-577X}
}

\authorrunning{D. Cambrin et al.}

\institute{Politecnico di Torino, Torino Italy \\
\email{\{daniele.regecambrin,paolo.garza\}@polito.it}\\
\and
University of Texas at San Antonio, San Antonio, TX USA\\
\email{\{isaac.corley,peyman.najafirad\}@utsa.edu}}

\maketitle

\begin{abstract}
Earthquakes are commonly estimated using physical seismic stations, however, due to the installation requirements and costs of these stations, global coverage quickly becomes impractical. An efficient and lower-cost alternative is to develop machine learning models to globally monitor earth observation data to pinpoint regions impacted by these natural disasters. However, due to the small amount of historically recorded earthquakes, this becomes a low-data regime problem requiring algorithmic improvements to achieve peak performance when learning to regress earthquake magnitude. In this paper, we propose to pose the estimation of earthquake magnitudes as a metric-learning problem, training models to not only estimate earthquake magnitude from Sentinel-1 satellite imagery but to additionally rank pairwise samples. Our experiments show at max a $30\%+$ improvement in MAE over prior regression-only based methods, particularly transformer-based architectures. Code, experimental results, and model checkpoints are openly available at \href{https://github.com/isaaccorley/earthquake-monitoring-by-ranking}{github.com/isaaccorley/earthquake-monitoring-by-ranking}.

\end{abstract}

\section{Introduction}

\begin{figure}[ht!]
\captionsetup[subfigure]{labelformat=empty}
\centering

\begin{subfigure}{0.32\linewidth}
\centering
\includegraphics[trim={0 0 0 0},clip,width=\textwidth]{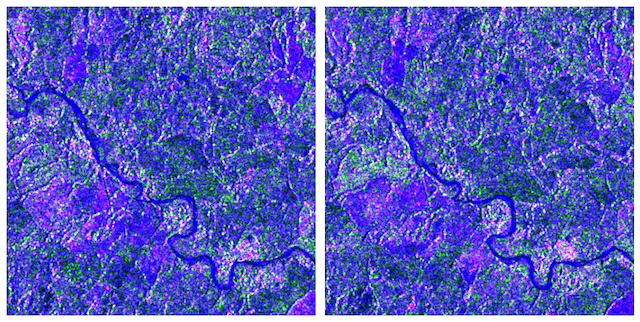}
\caption{Magnitude 5.14}
\end{subfigure}
\begin{subfigure}{0.32\linewidth}
\centering
\includegraphics[trim={0 0 0 0},clip,width=\textwidth]{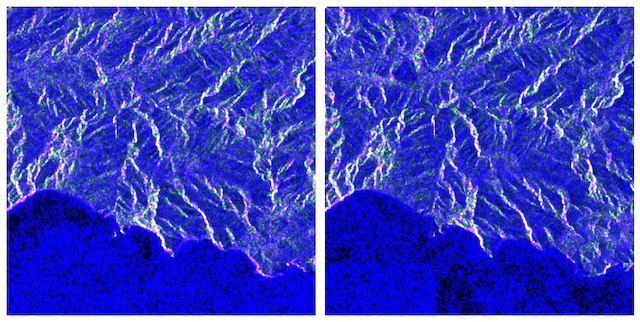}
\caption{Magnitude 5.20}
\end{subfigure}
\begin{subfigure}{0.32\linewidth}
\centering
\includegraphics[trim={0 0 0 0},clip,width=\textwidth]{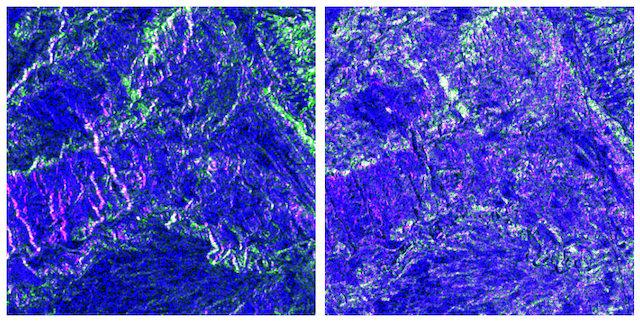}
\caption{Magnitude 5.58}
\end{subfigure}

\begin{subfigure}{0.32\linewidth}
\centering
\includegraphics[trim={0 0 0 0},clip,width=\textwidth]{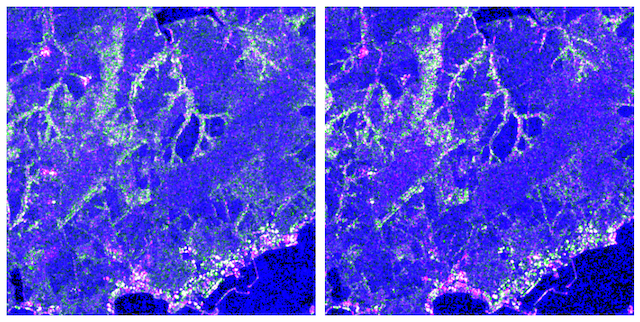}
\caption{Magnitude 5.61}
\end{subfigure}
\begin{subfigure}{0.32\linewidth}
\centering
\includegraphics[trim={0 0 0 0},clip,width=\textwidth]{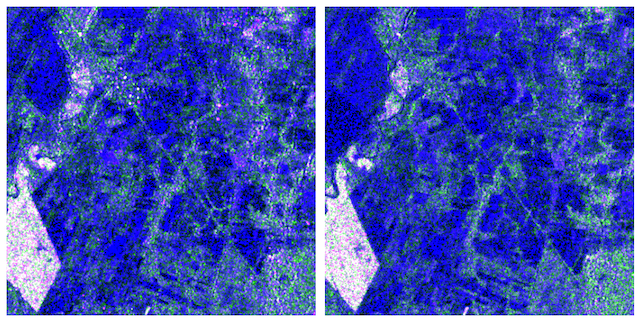}
\caption{Magnitude 6.09}
\end{subfigure}
\begin{subfigure}{0.32\linewidth}
\centering
\includegraphics[trim={0 0 0 0},clip,width=\textwidth]{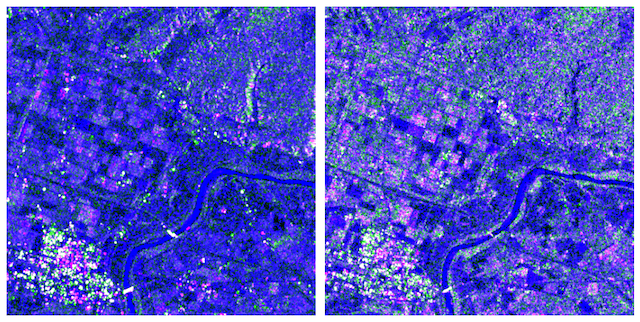}
\caption{Magnitude 6.48}
\end{subfigure}

\begin{subfigure}{0.32\linewidth}
\centering
\includegraphics[trim={0 0 0 0},clip,width=\textwidth]{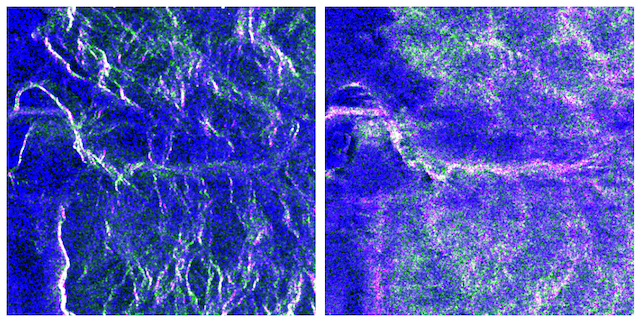}
\caption{Magnitude 6.54}
\end{subfigure}
\begin{subfigure}{0.32\linewidth}
\centering
\includegraphics[trim={0 0 0 0},clip,width=\textwidth]{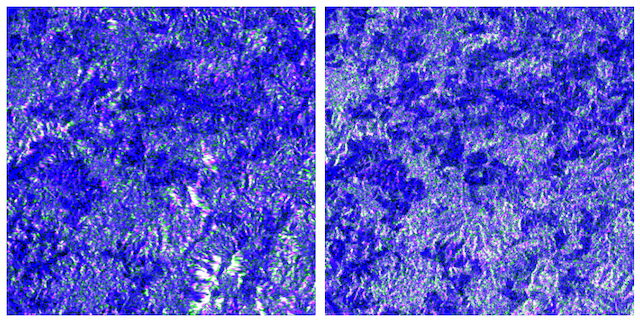}
\caption{Magnitude 6.56}
\end{subfigure}
\begin{subfigure}{0.32\linewidth}
\centering
\includegraphics[trim={0 0 0 0},clip,width=\textwidth]{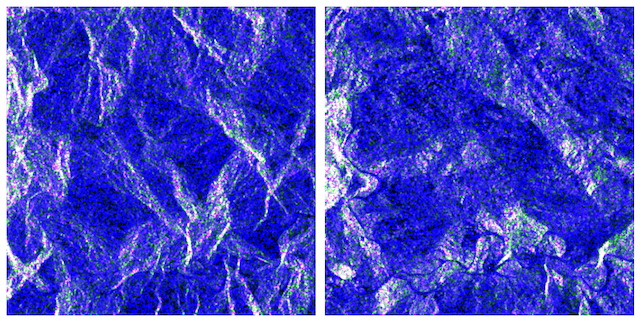}
\caption{Magnitude 6.69}
\end{subfigure}
\caption{\textbf{QuakeSet samples varied by earthquake magnitude}. Each sample contains a pair of pre and post earthquake event Sentinel-1 (SAR) imagery containing 2 bands (VV \& VH). The samples in this figure are plotted as false color images (VV, VH, VV/VH) along with their magnitudes.}
\label{fig:samples}
\end{figure}

Earthquakes are one of the most devastating natural disasters, causing significant loss of life, property, and economic damage~\cite{earthquakes_cost}. Accurate and timely detection and magnitude estimation of earthquakes are crucial for effective disaster response and mitigation. Traditionally, this task relies on an extensive network of physical seismic stations~\cite{earthquake_transformers}. However, the installation and maintenance of these stations are resource-intensive and logistically challenging, limiting their global coverage.

With the advent of satellite technology, earth observation data has become increasingly accessible~\cite{sentinel1}, providing a promising alternative for monitoring seismic activity. Leveraging this data, machine learning models can be developed to detect and estimate the magnitude of earthquakes on a global scale~\cite{quakeset}. Despite this potential, the application suffers from the limited amount of historical earthquake data combined with satellite data, which poses a low-data regime problem that could limit the performance of regression models.

To overcome this challenge, we propose a novel approach that redefines the estimation of earthquake magnitudes as a metric-learning problem. Instead of solely focusing on regression, our method trains models to not only predict earthquake magnitudes from Sentinel-1 satellite imagery, but also to rank pairwise samples based on their relative magnitudes. This dual-objective training enhances the model's ability to generalize from limited data, improving its accuracy and robustness.

Our contributions are twofold and can be summarized as follows:

\vspace{-4mm}

\begin{enumerate}
    \item We formulate earthquake magnitude estimation as a metric-learning problem, incorporating both regression and ranking objectives to facilitate learning.
    \item We apply this solution to several state-of-the-art convolutional and transformer based models, demonstrating a significant improvement in Mean Absolute Error (MAE) of over 30\%.
\end{enumerate} 

\section{Related Works}

\paragraph{\textbf{\textup{Remote Sensing with SAR imagery}}}
Synthetic Aperture Radar (SAR) remote sensing has emerged as a critical technology for Earth observation, offering unique capabilities to capture high-resolution images under all-weather conditions, day or night. 
SAR imaging techniques have significantly improved, enhancing the data's quality and applicability. 
Due to its effectiveness under many weather conditions, SAR imagery was a choice for topographic mapping~\cite{copernicus_dem} and detecting ground deformation with high precision~\cite{sentinel1_deformation}. In agriculture, the use of SAR to monitor crop growth and soil moisture levels proves to be critical for precision farming~\cite{review_precision_agriculture}. In hazard management, it proved effective in many use cases like analyzing cyclones~\cite{sentinel1_cyclones}, floods~\cite{sen1floods}, and wildfires~\cite{sentinel1_wildfires}.

\paragraph{\textbf{\textup{Machine Learning in Seismology}}}

Interest in leveraging machine learning and deep learning techniques in seismology is growing~\cite{magnitude_estimation_review}. Machine learning methods are mainly employed for real-time earthquake detection through seismic waves (P-waves and S-waves). Additionally, these techniques help distinguish between earthquakes and microtremors~\cite{earthquake_microtremors}, and are employed in phase picking~\cite{earthquake_transformers,deep_learning_pwaves}. Current approaches include the use of convolutional neural networks~\cite{p_detector,deep_learning_pwaves}, generative adversarial networks~\cite{earthquake_microtremors}, and transformers~\cite{earthquake_transformers} directly on seismic wave data.
While Sentinel-1 imagery has been explored for earthquake analysis, it traditionally involved manual analysis rather than automated systems~\cite{sentinel1_earthquakes}. However, recent research by Cambrin et al.~\cite{quakeset} introduced the QuakeSet dataset, which uses bi-temporal Sentinel-1 imagery for earthquake detection, demonstrating the potential for automated solutions with SAR imagery.

\section{Methodology}

\paragraph{\textbf{\textup{Magnitude Estimation}}}
The magnitude estimation task can be formulated as follows. 
We have a set of time-series $\mathcal{T}$, where each time-series $T_s\in \mathcal{T}$ is composed of $N$ images, with $N = 2$, of size $W \times H \times C$ related to the same spatial area $S$ at different timesteps $\{T_1,\dots,T_N\}$ and a ground truth value $G_t \in [0...M_m]$ (where $M_m$ is the maximum value for the given magnitude scale). Given a training dataset $D_{tr}$, composed of a set of pairs $(T_s, G_t)$, we train a machine learning model $M$ to regress the magnitude $G_t$. Given a test dataset $D_{ts}$, composed of a set of $T_s$, we can regress $G_t$ for each sample $T_s$ using $M$.

\paragraph{\textbf{\textup{Margin Ranking Loss}}}
We employed Margin Ranking loss $L_{MR}$ to encourage the model to distinguish between high-magnitude samples and lower-magnitude ones. The loss is expressed in the following way:
\begin{align}
    L_{MR}(x_1,x_2,y)=max(0,-y*(x_1-x_2)+m) \\
    y_i = 
    \begin{cases}
    1 & \text{if } x_{1i} >= x_{2i}  \\
    -1 & \text{if } x_{1i} < x_{2i}
    \end{cases}
\end{align}
where $x_1$ is a vector of predicted magnitudes for a set $S_1$ of images, $x_2$ are the predicted magnitudes for a set $S_2$ and $m$ is the margin. $y$ is a vector (of elements $y_i$) which assume values $1$ or $-1$ based on the ranking between samples from $x_1$ and $x_2$. 
The final loss is expressed as the sum of the Mean Squared Error (MSE) loss, $L_{MSE}$, and Margin Ranking loss, $L_{MR}$:  
\begin{equation}
    L = L_{MSE} + L_{MR}
\end{equation}

\section{Experiments}

\paragraph{\textbf{\textup{Models}}}
To have a more complete overview of the possible architectures, we train a mixture of convolutional (ConvNeXt v2\cite{convnext} and MobileNet\cite{mobilenet}) and transformer-based (MobileViT v2\cite{mobilevit} and ViT-tiny\cite{vit}) models. This selection is further motivated to establish if model size is important in estimating earthquake magnitude.

\paragraph{\textbf{\textup{Experimental Settings}}}
Our experiments utilize the TorchGeo~\cite{stewart2022torchgeo} implementation of the QuakeSet dataset~\cite{quakeset} which contains 3,327 pairs of 2-channel Sentinel-1 imagery (VV and VH polarizations) of size $512\times512$ sampled at different timesteps, $T_1$ and $T_2$, before and after an earthquake, respectively. %

The models were trained with the AdamW optimizer~\cite{loshchilov2017fixing} for 10 epochs and a batch size of 16. We employed a linear learning rate scheduler with a warmup of 10\% of the total training steps and a peak learning rate of $\alpha=1e-4$. All models are initialized using ImageNet~\cite{deng2009imagenet} pretrained weights. We preprocess the raw Sentinel-1 imagery by converting to decibels and use random horizontal and vertical flip augmentations with a probability of 0.5. We set ranking loss margin $m$ to 0.02. Each image pair was concatenated along the channel axis to get samples of size $4 \times 512 \times 512$.

\paragraph{\textbf{\textup{Results}}}

\begin{table}[t!]
    \centering
    \caption{\textbf{Results on the QuakeSet dataset for Earthquake Magnitude Regression with ($\checkmark$) and without margin ranking loss.} Adding a margin ranking objective significantly improves the performance across all architectures.}
    \label{tab:regression_results}
    \vspace{2mm}
    \begin{tabular}{@{}c|cc|cc@{}}
\toprule
Model                        & \# Params $\downarrow$ & MFLOPs $\downarrow$       & $L_{MR}$ & MAE $\downarrow$   \\ \midrule
\multirow{2}{*}{ConvNeXtv2 - Atto}    & \multirow{2}{*}{3.4M}  & \multirow{2}{*}{373.08} &         & 0.2577    \\
                             &                        &                           & $\checkmark$        & \textbf{0.2287} \\ \midrule
\multirow{2}{*}{MobileNet-S} & \multirow{2}{*}{1.5M}  & \multirow{2}{*}{44.10}         &         & 0.4540  \\
                             &                        &                           & $\checkmark$        & \underline{0.3678}            \\ \midrule
\multirow{2}{*}{MobileNet-L} & \multirow{2}{*}{4.2M}  & \multirow{2}{*}{155.33}         &         & 0.5465  \\
                             &                        &                           & $\checkmark$        & \underline{0.5414}            \\ \midrule
\multirow{2}{*}{MobileViTv2-100}   & \multirow{2}{*}{4.4M}  & \multirow{2}{*}{964.60} &         & 0.3539 \\
                             &                        &                           & $\checkmark$        & \underline{0.3383}             \\ \midrule
\multirow{2}{*}{ViT-T}         & \multirow{2}{*}{5.7M}  & \multirow{2}{*}{1541.56}         &          & 0.3150 \\
                             &                        &                           & $\checkmark$         &  \underline{0.2613}            \\ \bottomrule
\end{tabular}
\vspace{-2mm}
\end{table}

We analyzed the models trained with $L_{MR}$ compared to models trained only with $L_{MSE}$ to understand the contribution of the additional term. We compare performance using the Mean Absolute Error (MAE) between the estimated and ground truth magnitudes as well as the FLOPS computed on an Intel(R) Core(TM) i9-10980XE CPU. 

In \Cref{tab:regression_results}, we can see this approach proves to be effective in any case, both with CNNs and ViTs. ConvNeXt provides the best results overall, particularly employing $L_{MR}$, followed by ViT. Although, it is still worse than ConvNeXt without $L_{MR}$. ViT attention increases resource consumption, which makes the network less competitive than a convolutional neural network.

MobileViT, thanks to the hybrid approach, provides good results with less FLOPS than ViT. Although this solution is still not competitive with convolutional solutions. MobileNet-Small (Mobilenet-S) proves to be a competitive solution by achieving a low MAE and the least amount of FLOPS. In comparison, MobileNet-Large (MobileNet-L) instead tends to overfit, obtaining the worst performance and margin loss, $L_{MR}$.

\section{Conclusion}
In this paper, we modified the task of learning to regress earthquake magnitudes from satellite imagery as a ranking problem. Our method benefits from requiring no additional data other than the existing magnitude labels, proving that the rankings between the predictions inside a batch can be employed to better add distance between samples with different magnitudes.

\vspace{-2mm}

\bibliographystyle{splncs04}
\bibliography{bibliography}
\end{document}